# Multi-task Cross-modal Learning for Chest X-ray Image Retrieval

Zhaohui Liang
Division of Intramural Research,
National Library of Medicine,
NIH
Bethesda, MD, USA
zhaohui.liang@nih.gov

Sivaramakrishnan Rajaraman
Division of Intramural Research,
National Library of Medicine,
NIH
Bethesda, MD, USA
sivaramakrishnan.rajaraman@nih.gov

Niccolo Marini
Division of Intramural Research,
National Library of Medicine,
NIH
Bethesda, MD, USA
niccolo.marini@nih.gov

Zhiyun Xue
Division of Intramural Research,
National Library of Medicine,
NIH
Bethesda, MD, USA
zhiyun.xue@nih.gov

Sameer Antani
Division of Intramural Research,
NIH
Bethesda, MD, USA
santani@mail.nih.gov

*Abstract*—CLIP and BiomedCLIP are examples of vision-language foundation models and offer strong cross-modal embeddings; however, they are not optimized for fine-grained medical retrieval tasks, such as retrieving clinically relevant radiology reports using chest X-ray (CXR) image queries. To address this shortcoming, we propose a multi-task learning framework to fine-tune BiomedCLIP and evaluate improvements to CXR image-text retrieval. Using BiomedCLIP as the backbone, we incorporate a lightweight MLP projector head trained with a multi-task composite loss function that includes: (1) a binary cross-entropy loss to distinguish normal from abnormal CXR studies, (2) a supervised contrastive loss to reinforce intra-class consistency, and (3) a CLIP loss to maintain cross-modal alignment. Experimental results demonstrate that the fine-tuned model achieves more balanced and clinically meaningful performance across both image-to-text and text-to-image retrieval tasks compared to the pretrained BiomedCLIP and general-purpose CLIP models. Furthermore, t-SNE visualizations reveal clearer semantic clustering of normal and abnormal cases, demonstrating the model's enhanced diagnostic sensitivity. These findings highlight the value of domain-adaptive, multi-task learning for advancing cross-modal retrieval in biomedical applications.

## I. Introduction

The success of large language model (LLM) driven artificial intelligence (AI) has created new opportunities to advance biomedical research and healthcare technology [1]. Medical domain-specific LLMs show strong potential to improve multimodal medical knowledge management and innovate clinical information retrieval. The Contrastive Language–Image Pretraining (CLIP) based models provide an effective approach of encoding both medical images and corresponding text (such as radiology reports) into a shared embedding space, enabling powerful cross-modal retrieval strategies. This capability is especially valuable in linking medical images such as X-ray images and computed tomography (CT) images with associated textual findings from massive medical databases, which allows clinicians and researchers to efficiently access similar cases, compare visual and textual patterns, and streamline diagnostic workflows. Furthermore, high-precision cross-modal retrieval of highly relevant radiology cases lays the foundation for downstream applications such as automated diagnosis and dynamic, context-aware radiology report generation. Unfortunately, the general-purpose vision-language models often lack the perfect alignment required for the specificity and complexity of medical data. Domain-adapted retrieval methods are therefore critical for bridging this gap, supporting more accurate and semantically meaningful cross-modal understanding in medical imaging.

Significant efforts have been devoted to cross-modal learning for medical data, particularly in radiology image analysis using various deep neural network architectures. For example, a framework named Pretext-Invariant Representation Learning (PIRL) was proposed as a self-supervised method to learn cross-modal representations through a dual-encoder model with adaptive gated fusion, aiming to improve image-text retrieval performance [2]. However, this approach struggles to capture subtle distinctions in visually similar images and is incapable of addressing domain-specific challenges critical for medical image diagnosis. Another cross-modal representation method, called MetaTrON, was introduced to encode radiology images and clinical reports into a shared embedding space using omnimodal fusion [3]. It combines vision-language pretraining with meta-learning strategies to enable robust and adaptive radiographic image triage, addressing the challenge of label scarcity in clinical radiology data. Despite its strengths, the method is constrained by computationally intensive meta-training loops and the need for manual pretraining of modality-specific encoders, which ultimately limits its scalability and applicability in real-time clinical settings. Similarly, the Bootstrapping Language-Image Pre-training models (BLIP and BLIP-2) leverage large-scale vision-language pretraining frameworks designed for general-domain image-text understanding and generation [4-5]. While effective for broad scope tasks, they rely heavily on large training datasets and fail to capture the fine-grained visual patterns essential in medical images, such as differentiating between normal and abnormal chest X-rays (CXR), which limits their utility in specialized clinical domains.

The CLIP model presents a more effective architecture for image-language cross-modal processing and overcomes the above limitations. Its dual-encoder design trained with a contrastive learning loss directly optimizes image-text alignment and enables learning of subtle pathogenic patterns in medical images and aiding clinical diagnostic tasks. Unlike PIRL or BLIP, which rely on reconstruction or generative losses, CLIP learns discriminative features by contrasting matched and unmatched image-text pairs. This learning mode makes it adept at handling subtle variations in radiology images. In addition, CLIP is more computationally efficient and supports scalable and independent encoding of image and text, which facilitates real-time retrieval use cases. All these characteristics make CLIP a compelling foundation model for domain-adapted, retrieval-focused medical applications [6]. Based on the strengths of CLIP, BiomedCLIP extends the CLIP architecture to the biomedical domain by leveraging a dual-encoder design that includes a Vision Transformer (ViT) for image encoding and a PubMedBERT-based encoder for biomedical text. This enables BiomedCLIP to learn semantically rich and domain-specific cross-modal embeddings from large-scale biomedical image-text pairs [7]. Therefore, hypothesize that fine-tuning BiomedCLIP on downstream datasets, such as CXR images paired with radiology reports, can better adapt its pretrained representations to the specific semantics, label distributions, and subtle diagnostic patterns of the target domain.

However, in our preliminary experiments, we observed that fine-tuning BiomedCLIP using a single-task objective, such as the original CLIP contrastive loss, did not yield performance gains and, in some cases, even degraded downstream task performance. This suggests that naively applying the same loss used during pretraining can result in suboptimal domain adaptation and potentially lead to catastrophic forgetting of pretrained knowledge, particularly for datasets with significant feature variety. This limitation is especially pronounced in the medical domain, where interpreting CXRs poses unique challenges. As highlighted by Armato et al. [8], detecting radiographic patterns with clinical or pathogenic meaning is inherently difficult due to subtle variations in contrast, shape, or texture that may resemble normal anatomical structures. As a result, disease indicators in CXR are often spatially diffuse, context-dependent, or overlap with normal anatomy, making them easily overlooked by visual models (e.g. convolutional neural network (CNN)) not explicitly trained for such fine-grained distinctions.

To address these challenges, we adopt a multi-task learning strategy that jointly optimizes for multiple objectives: a binary cross entropy loss, a supervised contrastive loss, and a CLIP-style contrastive loss. This strategy helps retain the generalizable representations learned during pretraining while enabling the model to learn task-specific features during fine-tuning. Hervella et al. showed that such multi-task optimization improves neural network training dynamics through shared representations [9], and a recent review on multi-task learning in medical imaging further supports its effectiveness in unifying related diagnostic tasks for enhanced performance [10].

Motivated by these findings, we propose a multi-task fine-tuning framework to improve the cross-modal retrieval performance of the BiomedCLIP model on CXR images and their corresponding radiology reports. Our approach adopts the pretrained BiomedCLIP as the backbone and incorporates a lightweight MLP projection head, trained using a composite loss that combines binary classification loss, supervised contrastive loss, and CLIP-style image-text contrastive loss. This multi-task objective is designed to jointly enhance abnormality discrimination, intra-class embedding cohesion, and cross-modal semantic alignment. We believe this approach offers a valuable direction for domain adaptation of large-scale vision-language models in specialized medical applications.

## II. METHODS

### A. Cross-modal Architecture

Our cross-modal model leverages the pretrained BiomedCLIP as the backbone, where all the pretrained weights were frozen except for the last layers of the image and text encoders. This design retains the domain-specific knowledge acquired during pretraining while allowing limited adaptation to the downstream tasks. The model architecture consists of a lightweight binary classification head built on top of the BiomedCLIP outputs. Specifically, input chest X-ray images are encoded using the Vision Transformer (ViT) branch, and the paired radiology report text is encoded using the PubMedBERT branch. These two encoders produce modality-specific embeddings, which are projected into a shared latent space. The resulting image and text embeddings (each of dimension 512) are fused via simple averaging to capture joint cross-modal semantics. The fused vector is then passed through a dropout layer for regularization, followed by a trainable linear projection (MLP head) that outputs a single binary logit for classifying normal vs. abnormal findings. This architecture ensures efficient fine-tuning while preserving the pretrained model's generalizable representations. The model architecture is illustrated in Fig. 1.

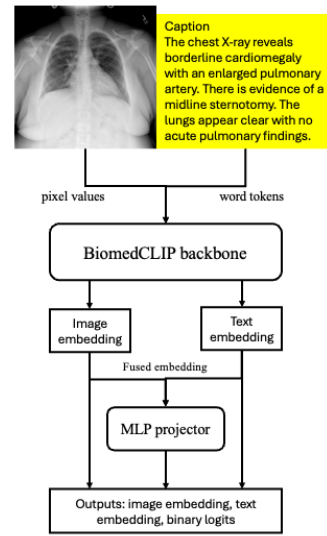

Fig. 1. Architecture of BiomedCLIP based model for multi-task optimization

## B. Multi-task learning with composite loss

Multi-task learning (MTL) is a paradigm in which a model is trained to perform multiple related tasks simultaneously to improve generalization by leveraging shared representations. Compared to single-task learning, where the model focuses on a single objective, MTL jointly optimizes multiple loss functions to capture complementary information across tasks. This approach is particularly beneficial in medical imaging, where supervision from diverse sources can lead to more robust and semantically meaningful feature representations. In our framework, we adopted a composite learning objective under the MTL setting to train the model not only to align CXR images with their corresponding radiology reports, but also to learn discriminative features for abnormality classification and to enhance intra-class feature consistency..

The original CLIP objective is a symmetric cross-modal contrastive loss that aligns image and text representations in a shared embedding space [6]. We applied L2 normalization to both image and text embedding so that cosine similarity can be computed via dot product. For each training batch, image-to-text and text-to-image logits are calculated using scaled dot products, with a temperature parameter $\tau$ to control the sharpness of the similarity distribution. Cross-entropy loss is applied to the softmax-normalized logits in both directions, and their average forms the final CLIP loss. This encourages high similarity for matching pairs, reinforcing semantic alignment across modalities. The CLIP loss is defined in Eq (1).

$$\mathcal{L}_{CLIP} = \frac{1}{2N}\sum_{i=1}^{N}\left[-\log\frac{\exp\left(\frac{v_i^T t_i}{\tau}\right)}{\sum_{j=1}^{N}\exp\left(\frac{v_i^T t_j}{\tau}\right)} - \log\frac{\exp\left(\frac{t_i^T v_i}{\tau}\right)}{\sum_{j=1}^{N}\exp\left(\frac{t_i^T v_j}{\tau}\right)}\right] \quad (1)$$

where $v_i \in R^D$ denotes the normalized image embedding for sample $i$; $t_i \in R^D$ denotes normalized text embedding for sample $i$, $\tau$ is the temperature hyperparameter, and $N$ is the batch size.

To further enhance image-text alignment, we incorporate a supervised contrastive loss as part of the training objective. Supervised contrastive learning extends contrastive methods by using label information to define positive pairs [11]. Unlike self-supervised contrastive loss, which contrasts each sample with a single positive [12], the supervised version (SupCon) considers all same-class samples in the batch as positives. The SupCon loss is defined in Eq (2).

$$\mathcal{L}_{SupCon} = \frac{1}{N}\sum_{i=1}^{N}\frac{1}{|P(i)|}\sum_{p\in P(i)} -\log\frac{\exp(f_i^T f_p/\tau)}{\sum_{a=1,a\neq i}^{N}\exp(f_i^T f_a/\tau)} \quad (2)$$

where $f_i \in R^D$ is the L2 normalized feature vector of sample $i$, $P(i)$ is the set of positive samples (i.e. samples with the identical class label as $i$), $\tau$ is the temperature hyperparameter controlling the sharpness of the similarity distribution.

This loss mathematically improves intra-class compactness and inter-class separability, enhancing the discriminative power of learned representations. It complements the CLIP objective by enforcing semantic consistency within the same class, rather than just across modalities—which is especially valuable in medical imaging, where subtle visual differences distinguish different medical diagnoses.

In medical image analysis, distinguishing abnormal images (with any disease) from normal images (no disease) is critical in medical diagnosis. To support this, we incorporated a binary cross-entropy (BCE) loss into the composite objective. BCE directly supervises binary classification and guides the model to learn clinically meaningful features that separate normal from abnormal CXRs. Unlike contrastive losses focused on embedding similarity, BCE captures diagnostic boundaries, and improves sensitivity to subtle or rare abnormalities. Therefore, including the BCE loss (Eq. 3) strengthens medical utility while preserving cross-modal alignment.

$$\mathcal{L}_{binary} = \frac{1}{N}\sum_{i=1}^{N}\left[-y_i \cdot \log\sigma(\hat{y}_i) - (1-y_i) \cdot \log(1-\sigma(\hat{y}_i))\right] \quad (3)$$

The total multi-task composite loss is formulated as a weighted sum of three components: binary cross-entropy (BCE) loss, supervised contrastive (SupCon) loss, and CLIP loss, each scaled by a corresponding coefficient $\lambda$:

$$\mathcal{L}_{total} = \lambda_1 \cdot \mathcal{L}_{binary} + \lambda_2 \cdot \mathcal{L}_{SupCon} + \lambda_3 \cdot \mathcal{L}_{CLIP} \quad (4)$$

Here, $\mathcal{L}_{binary}$ guides the binary classification task to distinguish normal from abnormal CXRs; $\mathcal{L}_{SupCon}$ promotes clustering of embeddings from the same class to improve intra-class consistency; and $\mathcal{L}_{CLIP}$ enforces alignment between CXR images and their corresponding captions, preserving cross-modal semantic coherence.

## III. DATASET AND EXPERIMENTS

### A. Dataset and Preprossing

To fine-tune the BiomedCLIP-based cross-modal retrieval model, we used the OpenI dataset from the U.S. National Library of Medicine [13]. It comprises 3,955 de-identified chest X-ray (CXR) radiology studies and 7,470 associated CXR images. Each study includes at least one posterior–anterior (PA) and one lateral view [14]. We used only the 3,955 PA view images, as they are the most clinically informative. The paired reports provided in XML format, were parsed to extract the "Findings" and "Impression" sections, which summarize key observations and diagnoses [15]. Approximately 38% of the reports are labeled normal and 62% as abnormal, which were used as binary labels for supervised learning. We reserved 400 image-text pairs (200 normal and 200 abnormal) for independent testing, with the remaining data split 90% / 10% for training and validation.

### B. Experiment Environment

The cross-modal retrieval model was implemented in Python and trained on an NIH Biowulf cluster node with an NVIDIA A100 GPU [16]. The training used a batch size of 128 and a multi-task composite loss combining binary cross entropy, supervised contrastive, and CLIP losses. We used a Bayesian optimization strategy to determine the optimal values of the loss weights $\lambda_1$, $\lambda_2$, and $\lambda_3$, searching from the empirical search space $\lambda_1 \in [0.1, 1]$, $\lambda_2 \in [0.2, 2]$, and $\lambda_2 \in [0.2, 2]$. After 20 trials, the best-performing combination was found to $\lambda_1 = 0.69$, $\lambda_2 = 1.97$, and $\lambda_3 = 0.46$. These optimized weights were used throughout training. The model built on the BiomedCLIP

This research is supported by the Division of Intramural Research of the National Library of Medicine, National Institutes of Health.

backbone was optimized with the AdamW optimizer using learning rates of $1 \times 10^{-5}$ for the BiomedCLIP backbone, and $1 \times 10^{-4}$ for the MLP projector. A two-stage learning scheduler was applied: 10% linear warm-up steps followed by cosine annealing for the remaining training steps over 20 epochs.

### C. Multi-task Cross-modal training

The multi-task cross-modal training algorithm extends the BiomedCLIP architecture by introducing a composite loss function aimed at enhancing the alignment between chest X-ray (CXR) images and their corresponding radiology report texts. Training begins with a dataset of paired CXR images and textual descriptions, each labeled as either normal or abnormal. To preserve the rich biomedical knowledge learned during pretraining and avoid suboptimal domain adaptation or catastrophic forgetting, we froze the entire BiomedCLIP backbone and only unfreeze the last layer of the image and text encoders. In each iteration, the model encodes both modalities into dense embeddings using these partially fine-tuned BiomedCLIP encoders. The resulting embeddings are fused and passed through a lightweight MLP projection head for binary classification. The training objective combines three loss components: (1) a binary cross-entropy (BCE) loss to supervise the classification of normal vs. abnormal CXRs, (2) a supervised contrastive (SupCon) loss that promotes tighter clustering of embeddings with the same label to improve intra-class consistency, and (3) a CLIP-style contrastive loss that reinforces semantic alignment between image and text embeddings. These components are integrated into a weighted composite loss, enabling the model to jointly optimize classification performance, representation quality, and cross-modal coherence within a unified framework. The full optimization procedure is illustrated in Fig. 2.

**Algorithm 1** Multi-task Cross-Modal Optimization Based on BiomedCLIP Architecture

**Require:** Paired image-caption dataset $\mathcal{D} = \{(I_i, T_i, y_i)\}_{i=1}^{N}$
1: Pre-trained BiomedCLIP model (trainable), classifier head $\text{Linear}(D, 1)$
2: Hyperparameters: learning rate $\alpha$, epochs $E$, batch size $B$, temperature $\tau$, weights $\lambda_1, \lambda_2, \lambda_3$
**Ensure:** Fine-tuned cross-modal classifier with learned alignment and representation
3: **for** epoch = 1 **to** $E$ **do**
4:    **for** each batch $(I, T, y) \in \mathcal{D}$ **do**
5:       $V \leftarrow \text{clip.encode\_image}(I)$     ▷ Image encoder
6:       $T_{\text{ids}} \leftarrow \text{Tokenize}(T)$
7:       $T_{\text{emb}} \leftarrow \text{clip.encode\_text}(T_{\text{ids}})$     ▷ Text encoder
8:       $H \leftarrow \text{Dropout}(\frac{V + T_{\text{emb}}}{2})$     ▷ Fused image-text features
9:       $\hat{y} \leftarrow \text{Linear}(H)$     ▷ Binary classification logits
10:      $V_{\text{norm}} \leftarrow \text{Normalize}(V)$
11:      $T_{\text{norm}} \leftarrow \text{Normalize}(T_{\text{emb}})$
12:      $F \leftarrow \frac{V_{\text{norm}} + T_{\text{norm}}}{2}$     ▷ Fused vector for SupCon
13:      $L_{\text{binary}} \leftarrow \text{BCEWithLogitsLoss}(\hat{y}, y)$
14:      $L_{\text{supcon}} \leftarrow \text{SupConLoss}(F, y)$
15:      $L_{\text{clip}} \leftarrow \text{CLIPLoss}(V, T_{\text{emb}})$
16:      $L_{\text{total}} \leftarrow \lambda_1 L_{\text{binary}} + \lambda_2 L_{\text{supcon}} + \lambda_3 L_{\text{clip}}$
17:      $\text{Backpropagate}(L_{\text{total}})$
18:      $\text{Update}(\text{all model parameters})$
19:      $\text{LogMetrics}(L_{\text{total}}, L_{\text{binary}}, L_{\text{supcon}}, L_{\text{clip}})$
20:    **end for**
21: **end for**
22: **return** Fine-tuned cross-modal model

Fig. 2. Algorithm of multi-task cross-modal optimization

### D. Cross-modal Vector Searching and Retrieval

We use Facebook AI Similarity Search (FAISS) with GPU acceleration for fast and scalable cross-modal retrieval of chest X-ray (CXR) images and associated reports. Fused embeddings obtained by averaging L2-normalized image and text embeddings are stored in a unified FAISS index to ensure modality-independent semantic representation. The fused embedding allows image or text queries to be matched in the same embedding space using cosine similarity. The top-k most similar results are retrieved along with metadata (e.g., image paths and labels), enabling efficient, semantically aligned search. This setup supports high-throughput clinical applications such as CXR case retrieval and report generation. An overview of the retrieval process is shown in Fig. 3.

**Algorithm 2** Cross-Modal Retrieval Based on Fused Vector Similarity

**Require:** Query vector $q \in \mathbb{R}^d$ (image or text)
1: Vector database of fused embeddings $I$
2: Image-text metadata $\mathcal{M}$
3: Number of top results $k$
**Ensure:** List of top-$k$ retrieved results with similarity scores
4: $q \leftarrow \frac{q}{\|q\|}$     ▷ Normalize the query embedding
5: $q_{\text{np}} \leftarrow \text{ConvertToNumpy}(q)$
6: $(S, \mathcal{I}) \leftarrow I.\text{search}(q_{\text{np}}, k)$     ▷ FAISS top-$k$ search
7: results $\leftarrow [\ ]$
8: **for** $j = 1$ to $k$ **do**
9:    idx $\leftarrow \mathcal{I}[0][j]$
10:   score $\leftarrow S[0][j]$
11:   meta $\leftarrow \mathcal{M}[\text{idx}]$     ▷ Get metadata
12:   results.append((meta, score))
13: **end for**
14: **return** results

Fig. 3. Algorithm of cross-modal retrieval

### E. Evaluation Metrics

We evaluate the fine-tuned cross-modal retrieval model trained with the proposed multi-task objective using two key metrics: retrieval accuracy and retrieval precision by the binary label (normal vs. abnormal CXR). Retrieval accuracy measures whether the correct paired item appears among the top-k most similar results in the shared embedding space, evaluated in both image-to-text (I→T) and text-to-image (T→I) directions. We report Top-1 and Top-k (k = 3, 5, 10) accuracy along with the average cosine similarity at each k. Results are compared against two benchmark models: the original pretrained but not domain-specific BiomedCLIP, and the general-purpose pretrained CLIP.

To assess clinical relevance, we also evaluate retrieval precision by the normal-abnormal binary labels, which check whether the top-k retrieved items share the same diagnostic category as the query. This captures higher-level semantic alignment beyond exact matches. Metrics include precision@k, F1 score, ROC AUC, and mean average precision (mAP) for both retrieval directions. This evaluation is essential for medical decision support, where semantically and diagnostically consistent retrieval enhances interpretability and utility.

## IV. RESULTS AND ANALYSIS

In this section, we report evaluation results for the domain-adapted BiomedCLIP model fine-tuned with multi-task learning. We first assess retrieval accuracy for image-to-text and text-to-image tasks. Then, we examine retrieval precision based on binary diagnostic labels to capture clinical relevance. Finally, we visualize the fused cross-modal embeddings using t-SNE to highlight the semantic separation between normal and abnormal chest X-rays.

## A. Cross-modal Retrieval Accuracy

We first evaluate retrieval accuracy, which measures how well the model retrieves the correct paired item—either a radiology report from a CXR image query or a CXR image from a report. Performance is assessed using top-k accuracy (k = 1, 3, 5, 10) and the average cosine similarity of the retrieved results.

TABLE I.  CROSS-MODAL RETRIEVAL ACCURACY

| Metrics | Fine-tuned model | BiomedCLIP | CLIP |
|---|---|---|---|
| *image-to-text retrieval* | | | |
| Accuracy@1 | 0.938 | **0.948** | 0.838 |
| Similarity score@1 | 0.845 | **0.845** | 0.819 |
| Accuracy@3 | 0.975 | **0.980** | 0.938 |
| Mean similarity score@3 | 0.832 | **0.832** | 0.814 |
| Accuracy@5 | **0.983** | **0.983** | 0.945 |
| Mean similarity score@5 | 0.827 | **0.827** | 0.813 |
| Accuracy@10 | **0.990** | 0.988 | 0.968 |
| Mean similarity score@10 | 0.821 | **0.822** | 0.811 |
| *text-to-image retrieval* | | | |
| Accuracy@1 | 0.630 | 0.563 | **0.940** |
| Similarity score@1 | **0.847** | 0.842 | 0.818 |
| Accuracy@3 | 0.790 | 0.695 | **0.988** |
| Mean similarity score@3 | **0.836** | 0.832 | 0.808 |
| Accuracy@5 | 0.845 | 0.763 | **1.000** |
| Mean similarity score@5 | **0.831** | 0.828 | 0.804 |
| Accuracy@10 | 0.925 | 0.855 | **1.000** |
| Mean similarity score@10 | **0.825** | 0.823 | 0.799 |

The fine-tuned BiomedCLIP model demonstrates balanced and robust performance across both image-to-text and text-to-image retrieval tasks. While the pre-trained BiomedCLIP slightly outperforms in image-to-text top-1 accuracy (0.9475 vs. 0.9375), the fine-tuned model maintains strong performance across top-k levels and achieves higher similarity scores in text-to-image retrieval. Unlike the CLIP model which excels only in text-to-image retrieval, our fine-tuned model consistently performs well in both directions. The result indicates improved cross-modal alignment and domain adaptation for chest X-ray retrieval. (TABLE I)

## B. Retrieval Precision by Binary Labels

Next, we evaluate the retrieval precision based on binary diagnostic labels: normal CXR vs abnormal CXR images. Retrieving category-consistent information is clinically critical, as it supports downstream tasks such as report generation and decision support grounded in diagnostic semantic similarity. The results are summarized in TABLE II & TABLE III.

TABLE II.  IMAGE-TO-TEXT RETRIEVAL PRECISION BY BINARY LABELS

| Metrics | Fine-tuned model | BiomedCLIP | CLIP |
|---|---|---|---|
| Precision@1 | 0.958 | **0.975** | 0.9050 |
| Similarity score@1 | 0.845 | **0.845** | 0.8186 |
| F1 score | 0.957 | **0.975** | 0.9112 |
| ROC AUC | 0.463 | 0.473 | **0.5134** |
| mAP | 0.500 | 0.506 | **0.5252** |
| Precision@3 | 0.993 | **1.000** | 0.9900 |
| Mean similarity score@3 | 0.832 | **0.832** | 0.8143 |
| Precision@5 | 0.998 | **1.000** | **1.0000** |
| Mean similarity score@5 | 0.827 | **0.827** | 0.8127 |
| Precision@10 | **1.000** | **1.000** | **1.0000** |
| Mean similarity score@10 | 0.821 | **0.822** | 0.8108 |

TABLE III.  TEXT-TO-IMAGE RETRIEVAL PRECISION BY BINARY LABELS

| Metrics | Fine-tuned model | BiomedCLIP | CLIP |
|---|---|---|---|
| Precision@1 | 0.950 | 0.933 | **1.000** |
| Similarity score@1 | **0.847** | 0.842 | 0.818 |
| F1 score | 0.949 | 0.931 | **1.000** |
| ROC AUC | 0.324 | 0.270 | **0.539** |
| mAP | 0.405 | 0.375 | **0.548** |
| Precision@3 | 0.983 | 0.978 | **1.000** |
| Mean similarity score@3 | **0.836** | 0.832 | 0.808 |
| Precision@5 | 0.998 | 0.985 | **1.000** |
| Mean similarity score@5 | **0.831** | 0.828 | 0.804 |
| Precision@10 | 0.998 | **1.000** | **1.000** |
| Mean similarity score@10 | **0.825** | 0.823 | 0.799 |

The retrieval precision results show that the fine-tuned BiomedCLIP model achieves strong and balanced performance across both image-to-text and text-to-image tasks. It consistently delivers high precision at top-k levels and competitive F1 scores and similarity metrics. While the general-purpose CLIP model achieves higher ROC AUC and mAP, reflecting broader embedding generalization. The fine-tuned BiomedCLIP demonstrates better alignment with diagnostic labels, indicating stronger clinical relevance. These results highlight the effectiveness of multi-task fine-tuning in improving semantic consistency and retrieval precision for medical applications.

## C. Cross-modal Embedding Cluster with t-SNE

The t-distributed stochastic neighbor embedding (t-SNE) is a dimensionality reduction technique that projects high-dimensional embeddings into a 2D space. It allows visual inspection of the embedding structure. Using t-SNE to visualize the embeddings of cross-modal models reveals how well the image and text embedding clusters on the diagnostic semantic labels (i.e., normal CXR studies vs abnormal CXR studies in our case). Well-separated clusters in the plot indicate strong alignment and discrimination between categories, reflecting the model's ability to learn meaningful and consistent cross-modal representations. (Fig. 4)

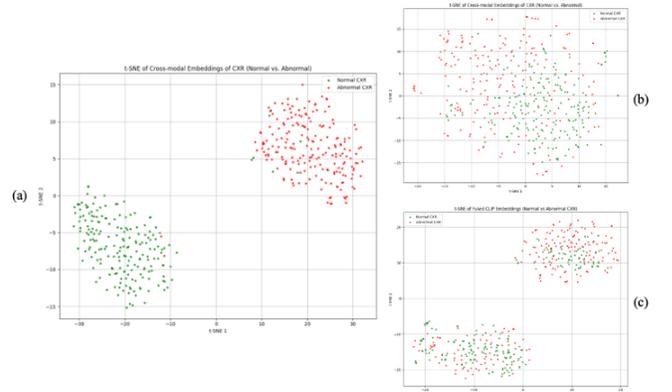

Fig. 4.  t-SNE cluster of cross-modal models
(a) fine-tuned model, (b) BiomedCLIP, (c) CLIP

The t-SNE visualizations highlight clear differences in how each model captures diagnostic semantics in the shared embedding space. The fine-tuned model exhibits strong clustering, with normal (green) and abnormal (red) CXR studies that form two well-separated groups with minimal overlap. This distinct separation suggests the model effectively learns clinically meaningful representations, and improves its utility for medical retrieval and decision support.

In contrast, the pretrained BiomedCLIP and general-purpose CLIP model show weaker clustering. BiomedCLIP embeddings are scattered and intermixed across labels, reflecting limited diagnostic alignment without task-specific supervision. CLIP demonstrates moderate separation, with some grouping by label, but notable overlap remains. The comparison highlights the importance of supervised fine-tuning in achieving reliable diagnostic discrimination for cross-modal biomedical tasks.

## V. Conclusion and Discussion

In this study, we introduced a multi-task fine-tuning framework for cross-modal biomedical retrieval using a domain-adapted BiomedCLIP model. By jointly optimizing binary classification, supervised contrastive learning, and CLIP loss, our approach improves semantic alignment between CXR images and the corresponding radiology reports. Experimental results demonstrate that the fine-tuned model achieves superior retrieval accuracy and diagnostic consistency, with well-separated clusters of normal and abnormal cases in the embedding space. Compared to the pretrained BiomedCLIP and the general-purpose CLIP, our new method yields more balanced and clinically relevant performance across both image-to-text and text-to-image cross-modal retrieval tasks.

These findings highlight the value of task-specific supervision and multi-task optimization in cross-modal biomedical retrieval. By integrating complementary learning signals, the CLIP based model effectively captures the subtle semantic and diagnostic features critical for clinical diagnosis and decision support. Nonetheless, the current framework focuses on binary classification, which simplifies the rich complexity of radiological findings. Future research should extend this method to multi-label classification of diverse pathologies, incorporate structured clinical metadata, and leverage larger, more heterogeneous datasets for greater generalizability. Our future research will focus on improving the multi-task learning optimization algorithms. Approaches such as reinforcement learning-based objective balancing, or algorithms based on collaborative Nash equilibrium learning, may offer principled strategies to harmonize competing losses and enhance training efficiency and stability.

In conclusion, our results demonstrate the effectiveness of multi-task fine-tuning in advancing cross-modal medical retrieval. The proposed framework outperforms both the domain-pretrained BiomedCLIP and the general-purpose CLIP model, demonstrating the importance of task-aware adaptation from foundation models in medical AI. This work lays a strong foundation for future clinical retrieval systems and motivates further research into more robust and theoretically grounded multi-task optimization strategies.

## Acknowledgment

This research was supported by the Intramural Research Program of the National Institutes of Health (NIH). The contributions of the NIH author(s) are considered Works of the United States Government. The findings and conclusions presented in this paper are those of the authors and do not necessarily reflect the views of the NIH or the U.S. Department of Health and Human Services. This work utilized the computational resources of the NIH HPC Biowulf cluster (https://hpc.nih.gov).